\documentclass[conference]{IEEEtran}

\usepackage{amsmath,amssymb,amsfonts}
\usepackage{algorithmic}
\usepackage{graphicx}
\usepackage{textcomp}
\usepackage{float}
\usepackage{xcolor}
\usepackage[style=ieee]{biblatex}
\newcommand{\fig}{Figure }
\newcommand{\eq}{Equation }

\newcommand{\loihiog}{†Workloads characterized on an Oheo Gulch system with a N3C2-revision Loihi 2 chip running an unreleased patch of NxCore 2.5.10 and NxKernel 0.4.0.}
\newcommand{\loihiqubo}{†Workloads characterized on a Kapoho Point system with eight N3D1-revision Loihi 2 chips running NxCore 2.5.18 and NxKernel 0.4.0.}

\def\BibTeX{{\rm B\kern-.05em{\sc i\kern-.025em b}\kern-.08em
    T\kern-.1667em\lower.7ex\hbox{E}\kern-.125emX}}
\begin{document}

\title{A Compute and Communication Runtime Model for Loihi 2\\
}

\author{\IEEEauthorblockN{Jonathan Timcheck}
\IEEEauthorblockA{\textit{Intel Corporation} \\
Santa Clara, CA, USA \\
jonathan.timcheck@intel.com}
\and
\IEEEauthorblockN{Alessandro Pierro}
\IEEEauthorblockA{\textit{LMU Munich, Intel Corporation} \\
Munich, Germany \\
alessandro.pierro@ifi.lmu.de}
\and
\IEEEauthorblockN{Sumit Bam Shrestha}
\IEEEauthorblockA{\textit{Intel Corporation} \\
Santa Clara, CA, USA \\
sumit.bam.shrestha@intel.com}
}

\maketitle

\begin{abstract}

Neuromorphic computers hold the potential to vastly improve the speed and efficiency of a wide range of computational kernels with their asynchronous, compute-memory co-located, spatially distributed, and scalable nature.
However, performance models that are simple yet sufficiently expressive to predict runtime on actual neuromorphic hardware are lacking, posing a challenge for researchers and developers who strive to design fast algorithms and kernels.
As breaking the memory bandwidth wall of conventional von-Neumann architectures is a primary neuromorphic advantage, modeling communication time is especially important.
At the same time, modeling communication time is difficult, as complex congestion patterns arise in a heavily-loaded Network-on-Chip. 
In this work, we introduce the first max-affine lower-bound runtime model---a multi-dimensional roofline model---for Intel's Loihi 2 neuromorphic chip that quantitatively accounts for both compute and communication based on a suite of microbenchmarks.
Despite being a lower-bound model, we observe a tight correspondence (Pearson correlation coefficient greater than or equal to 0.97) between our model's estimated runtime and the measured runtime on Loihi 2 for a neural network linear layer, i.e., matrix-vector multiplication, and for an example application, a Quadratic Unconstrained Binary Optimization solver. 
Furthermore, we derive analytical expressions for communication-bottlenecked runtime to study scalability of the linear layer, revealing an area-runtime tradeoff for different spatial workload configurations with linear to superlinear runtime scaling in layer size with a variety of constant factors. 
Our max-affine runtime model helps empower the design of high-speed algorithms and kernels for Loihi 2. 
\end{abstract}

\begin{IEEEkeywords}
neuromorphic, performance model, roofline
\end{IEEEkeywords}

\section{Introduction}

Neuromorphic computers such as Loihi 2 utilize computational primitives of the brain to deliver massive efficiency gains over conventional architectures across a diverse spectrum of tasks, ranging from optimization problems to artificial intelligence; however, mainstream adoption of neuromorphic hardware is slow \cite{muir2025road,schuman2022opportunities,
davies2021advancing,
meyer2025diagonal,
abreu2025neuromorphic,
pierro2025accelerating,
shrestha2024efficient}.
Despite impressive efficiency results, mainstream adoption is hindered due to neuromorphic hardware and software diversity and the deviation from the traditional von-Neumann architecture \cite{kudithipudi2025neuromorphic}.
These factors are especially significant when competing against incumbent architectures, which have heavily optimized software and large, established developer communities \cite{hooker2021hardware}.

In response, neuromorphic researchers have called for clear benchmarks on important real-world tasks to help facilitate progress toward ever-stronger evidence for neuromorphic supremacy \cite{schuman2022opportunities, timcheck2023intel,stewart2024focus,davies2019benchmarks}.
Such task-level or application-level benchmarks with straightforward evaluation criteria, such as accuracy, energy, and latency, hold the potential to bring clarity to comparisons among different neuromorphic architectures and algorithms \cite{yik2025neurobench}.

However, high-level benchmarks entangle hardware and algorithm innovation, which can impede hardware experts and algorithm experts from independently advancing their respective areas of expertise.
Well-designed abstractions facilitate harmonious development across technology stack layers---from software, AI frameworks, hardware, compilers, and so on. Indeed, these interfacing abstractions are routinely credited with accelerating development \cite{martin2009clean,patterson2016computer}.
In neuromorphic computing, while there certainly will always be an irreducible interdisciplinary component of algorithm-hardware co-design, the lack of a precise quantitative abstraction through which hardware and algorithms experts can readily exchange and build upon each other's innovations suggests that a layer of abstraction is missing.
For algorithm-hardware co-design, the missing layer of abstraction is a performance model, i.e., a simplified model of the hardware that quantitatively predicts the performance of an algorithm (runtime, energy, etc.).
Indeed, such performance models are the workhorse of kernel developers and hardware designers in mainstream architectures. 
For example, the roofline model for GPUs, which distinguishes between compute-bound and memory-bandwidth-bound kernels and provides a quantitative lower-bound on kernel runtime, is routinely used by kernel writers to optimize performance and by hardware designers to understand the impact of design tradeoffs \cite{williams2009roofline,yang2020hierarchical}.

Certainly, significant progress has been made toward neuromorphic performance models; however, a performance model that captures the primary aspects of the neuromorphic computing paradigm and is quantitatively predictive of performance on actual neuromorphic hardware remains elusive.
For instance, the algorithmic track of NeuroBench \cite{yik2025neurobench} utilizes the metrics memory footprint, connection sparsity, activation sparsity, and number of synaptic operations (SynOps).
These metrics capture neuromorphic principles to some degree---for instance, sparsity is often a good fit for neuromorphic hardware---however, these metrics do not have direct visibility into communication costs, which can account for significant time and energy expenditure, especially at scale \cite{mutlu2022modern}.
In the Intel Neuromorphic Deep Noise Suppression Challenge Algorithm Track \cite{timcheck2023intel}, a linear combination of SynOps and NeuronOps (neuron updates) was used as a proxy for energy.
This too includes two key neuromorphic computational costs and their relative importance; however, this again neglects energy expenditure from communication, and runtime is not modeled.
From a computational complexity perspective, \cite{aimone2025neuromorphic} defines a theoretical model for ideal and realized neuromorphic computers.
These models include many aspects of neuromorphic computation and are powerful tools for hardware-independent algorithm analysis; however, they can lack predictive power, as details of actual neuromorphic hardware become important, such as communication congestion.
\cite{yik2025modelingoptimizingperformancebottlenecks} presents a differential analysis of workload configurations resulting in a floor line runtime model showing implied compute, memory, and communication bottlenecks across three neuromorphic hardware platforms for a number of workloads with varying weight sparsity and activation sparsity. 
This analysis is helpful, as it can recommend the best direction in which to perform optimizations. However, its differential nature requires measuring runtime on neuromorphic hardware under several configurations to reveal the appropriate optimization direction, and it does not provide visibility into the specific causes of implied communication bottlenecks, leaving communication optimization difficult.

In this work, we introduce a max-affine (multi-dimensional roofline) runtime model for Intel's neuromorphic research chip, Loihi 2.
Our model takes into account compute operations, such as Synaptic Operations (SynOps), often known as Multiply-Accumulate Operations (MACs) in the deep learning literature, as well as communication on a potentially congested Network-on-Chip (NoC), and provides a lower-bound runtime estimate for an arbitrary algorithm.
We introduce a suite of simple microbenchmark workloads that run on Loihi 2 to calibrate our max-affine model, and we validate our model against measured runtime on Loihi 2 for a fundamental operation: a linear layer, i.e., vector-matrix multiplication. 

We find that, although our max-affine model produces a lower-bound on runtime, its predicted runtime is close to and highly correlated with measured runtime on Loihi 2.
This holds for a variety of linear layer shapes, sizes, and partitions, including those bottlenecked by compute, communication, or other factors. 
Importantly, we also find that our model's predicted runtime is highly correlated with measured runtime for sparse linear layers that stress the communication mesh on Loihi 2, across a variety of spatial placement patterns, further validating the more complex traffic resolution time term in the model.
We also demonstrate that our model is predictive of runtime for an example application, a Quadratic Unconstrained Binary Optimization (QUBO) solver \cite{pierro2024solving}.
Across all workloads measured in this paper, the Pearson correlation coefficient between measured and estimated runtime is $\geq$ 0.97.

Finally, we derive analytical expressions for the traffic resolution time in a linear layer, and we show an area-runtime tradeoff in which a naive rectangular spatial placement yields superlinear runtime scaling in layer size and that alternative spatial placements can yield linear time scaling with a variety of constant factors.
Our work can help algorithm designers and developers to conceptualize and write fast kernels on Loihi 2, as our model specifically addresses the difficult task of modeling runtime amid communication congestion. 

Notably, this work only models runtime, primarily utilizing the execution time profiling features of Loihi 2.
In addition to execution time profiling,  Loihi 2 features a full set of profiling capabilities including power and resource utilization.
The publication of a more comprehensive performance model for Loihi 2, including these additional aspects, is forthcoming.

\section{Profiling on Loihi 2}

Profiling on Loihi 2 requires an approach that matches its neuromorphic nature, so we begin with a general description of Loihi 2 systems and highlight the most salient components of a Loihi 2 chip.
Then we discuss the key performance profiling features available on Loihi 2 that are relevant to this work, such as SynOp counters, and we introduce a profiling tool called TrafficStats to gain insight into intrachip communication time in NoC traffic.

\subsubsection{Loihi 2 Systems}
Loihi 2 neuromorphic systems are composed of one or more Loihi 2 chips connected with an asynchronous low-latency event-driven communication mesh.
Loihi 2 systems are highly scalable; systems have been constructed with up to 1152 Loihi 2 chips, totaling 1.15 billion neurons 
\cite{intel_neuromorphic_system}.
In this work, however, we focus on a single Loihi 2 chip, as the essence and interplay of neuromorphic compute and communication are already critical to performance modeling at the single-chip scale. 
Please see \cite{Intel_Loihi2_2021} for a detailed description of Loihi 2. 

\begin{figure}[h]
\centerline{\includegraphics[width=1\columnwidth]{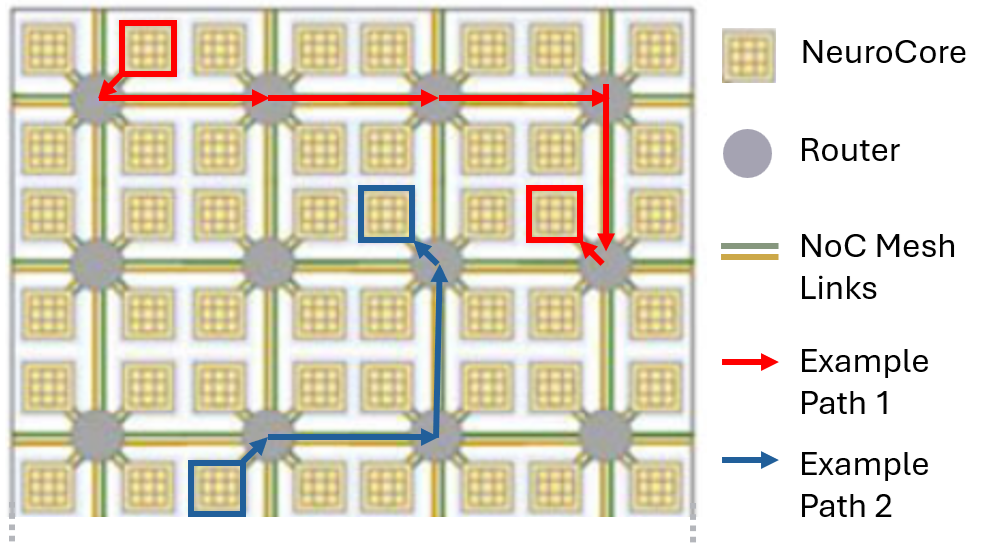}}
\vspace{-2mm}
\caption{Schematic illustrating the primary components of Loihi 2 (top half of chip shown):  NeuroCores, routers, and Network-on-Chip (NoC) mesh links.
NeuroCores perform computations, emit integer-valued messages known as graded spikes, and store synaptic connectivity matrices to weigh incoming graded spikes from other NeuroCores.
Routers direct messages from one NeuroCore to another along the NoC mesh links according to dimension-order routing: horizontally first, and then vertically.
Two example message paths are illustrated with red and blue arrows.
}
\label{fig:l2schematic}
\end{figure}

A single Loihi 2 chip is primarily composed of NeuroCores, Routers, and mesh links (\fig\ref{fig:l2schematic}). 
NeuroCores each have memory and compute: they store neuron states and synaptic weights, and they also perform computations, such as neuron updates and synaptic MAC operations.
Routers facilitate communication between NeuroCores. 
Via two bidirectional mesh links, each router is connected to four NeuroCores, and routers are connected to nearest-neighbor routers in the four cardinal directions.
Messages from NeuroCores are routed to their destinations according to dimension-order (X-Y) routing, as illustrated in \fig\ref{fig:l2schematic}.
The NoC also supports a synchronization operation, known as barrier synchronization, to synchronize algorithmic timesteps across NeuroCores: for many neuromorphic algorithms, e.g., a leak-integrate-and-fire network \cite{tal1997computing} evolving in discretized time, execution proceeds in a sequence of algorithmic timesteps, and it is critical that NeuroCores are synchronized to this degree. 
Other cores, such as CPU cores, are used for management or other special functions.
NeuroCores, routers, and mesh links handle the brunt of Loihi 2 workloads, so they are the focus of our runtime model.

\subsubsection{Loihi 2 Profiling Tools}
Loihi 2 has many built-in features to aid in profiling the operation of these components.
In this work, we are primarily concerned with measuring an algorithm's runtime, and Loihi 2 conveniently includes runtime probes for this purpose.
Furthermore, each NeuroCore also has hardware counters for various operations.
We can read these counters on each NeuroCore to understand how many of each operation was performed after a workload is executed.
Synaptic Operations (SynOps), Synaptic Memory Reads (SynMem reads), and Dendrite Operations (DendOps, also known as NeuronOps or neuron updates) are heavily utilized operations in neural networks run on Loihi 2, and so the hardware counters for these operations will be important inputs for the compute portion of our performance model.

Communication, however, cannot be inferred from simple hardware counters, as messages travel via various paths.
We therefore introduce a tool called TrafficStats which infers the bandwidth used on every mesh link.
TrafficStats inspects the configured NeuroCores and their message programs to infer how much message bandwidth will be loaded on every mesh link, taking into account dimension-order routing for each message.
TrafficStats assumes a uniform message-send frequency, i.e., that every neuron spikes on every algorithmic timestep, which is the case for the linear layer workloads we consider in this work.
However, TrafficStats can also be used to compute mesh link bandwidth loads for more specific activity patterns. E.g., we apply an activation sparsity factor to estimate traffic for the QUBO solver.
The TrafficStats tool is available in NxKernel, a flexible, low-level, and performant software package for mapping networks to Loihi 2.
Mesh link loads inferred from TrafficStats are visualized in the Loihi 2 traffic schematics in this paper (\fig\ref{fig:microbenchmarks_2}).

\section{Microbenchmark suite}

\begin{figure}[h]
\vspace{-0.1mm}
\centerline{\includegraphics[width=\columnwidth]{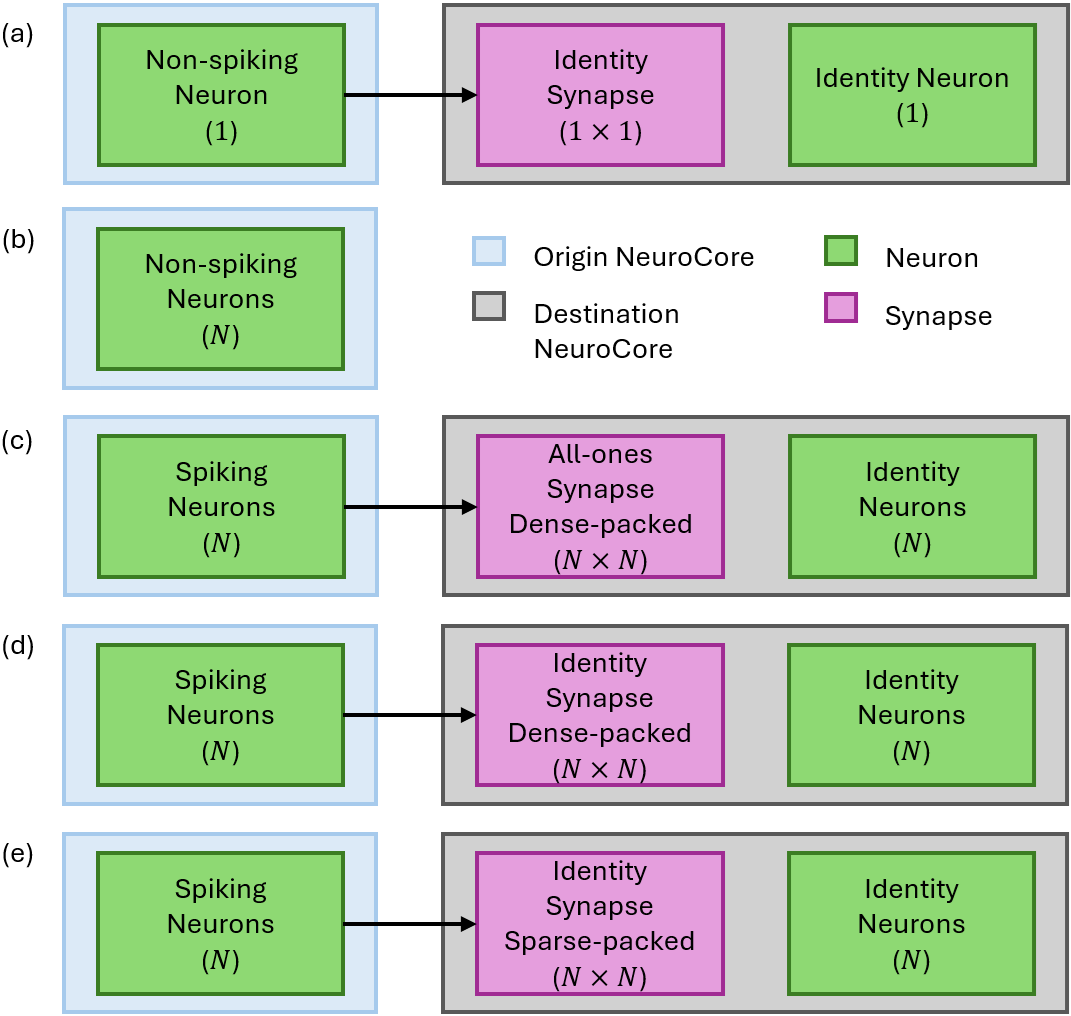}}
\vspace{-2mm}
\caption{Microbenchmark suite. (a) Barrier synchronization microbenchmark: dummy network with one sending and receiving neuron and no messages sent. Runtime is dominated by timestep coordination through the barrier synchronization mechanism. 
(b) DendOp microbenchmark: a single-core workload with many neurons that perform neuron updates (DendOps); DendOps dominate the processing time.
(c) SynOp microbenchmark: $N$ neurons send to $N$ neurons through a dense-packed all-ones synaptic weight matrix; a dominating $N^2$ SynOps are performed each timestep.
(d) SynMem read microbenchmark: a dense-packed identity synaptic connectivity matrix reduces SynOps to $N$ per timestep, while keeping a dominant $O(N^2)$ SynMem reads per timestep.
(e) Link bandwidth microbenchmark: $N$ messages are transferred from one core to another and the sparse-packed identity matrix connectivity results in few SynOps and SynMem reads. This workload is replicated to put heavy traffic along a common link (\fig\ref{fig:microbenchmarks_2}a), whose limited bandwidth dominates runtime. 
\vspace{-2mm}
}
\label{fig:microbenchmarks_1}
\end{figure}

\begin{figure}[h]
\centerline{\includegraphics[width=1\columnwidth]{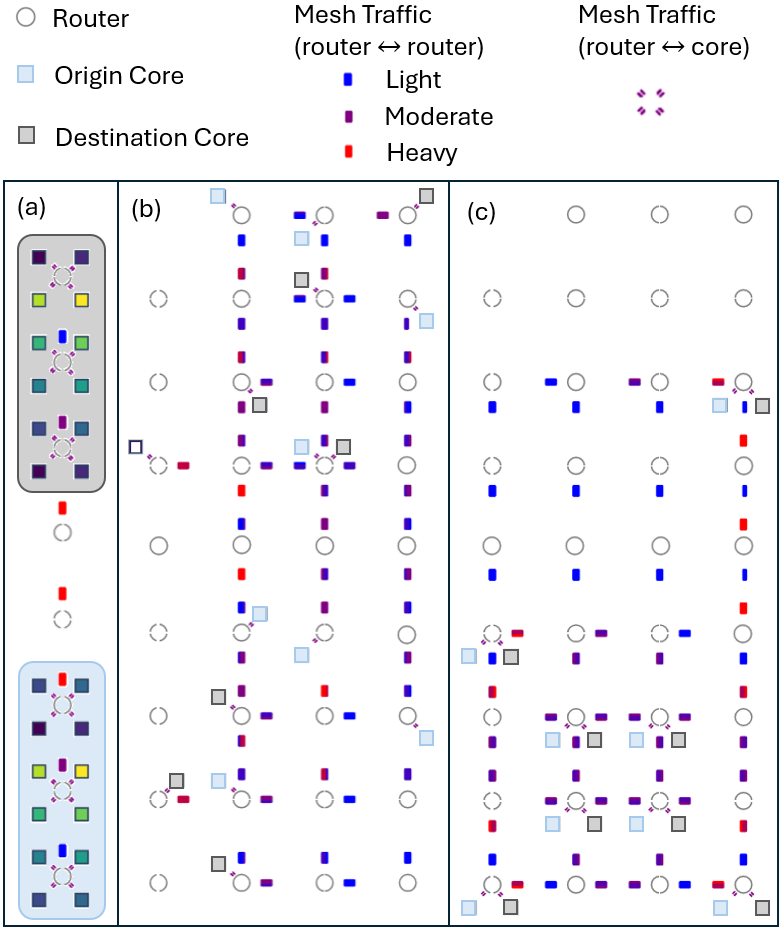}}
\vspace{-1mm}
\caption{NoC traffic visualizations. (a) Mesh link bandwidth microbenchmark. Several replicas (in this case 12) of the link bandwidth workload (\fig\ref{fig:microbenchmarks_1}e) are arranged in a single router column; this stresses the vertical link, allowing the peak link bandwidth to be measured.
(b) and (c) Two example NeuroCore placement patterns. The same 16 NeuroCores of the tiled-identity linear layer workload are mapped to different locations according to a (b) random placement or (c) X-shaped placement. 
Aggregate traffic emanating from each router is illustrated as a colored rectangle near the router, and the two halves of the rectangle represent the two meshes on Loihi 2;
coloring is relative to within each panel; the upper right corner of the X-shape is offset due to the default reservation of the displaced cores for special purposes.
\vspace{-4mm}
}
\label{fig:microbenchmarks_2}
\end{figure}

\begin{figure}[h]
\centerline{\includegraphics[width=\columnwidth]{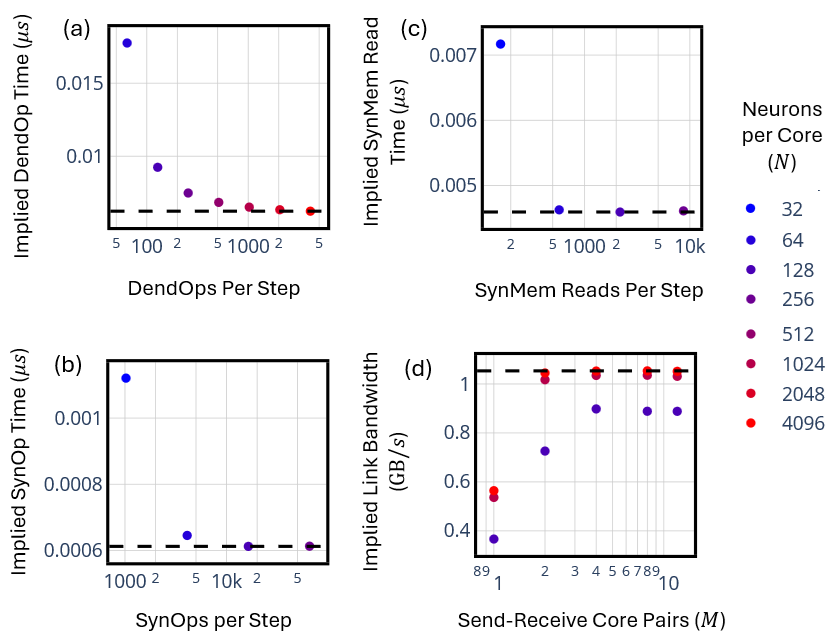}}
\vspace{-2.5mm}
\caption{
Microbenchmark measurements.
As each microbenchmark is scaled, the time per step asymptotes to a particular value representing peak system performance for the component stressed by the microbenchmark. This allows the implied effective (a) DendOp time, (b) SynOp time, (c) SynMem read time, and (d) link bandwidth to be calculated.
\loihiog
\vspace{-3mm}
}
\label{fig:microbenchmarks_measurements}
\end{figure}

In order to create a runtime model that is high-level enough to interface with algorithm designers and low-level enough to be predictive of actual hardware performance, we characterize the key compute and communication runtime bottlenecks on Loihi 2 with a suite of microbenchmarks.
Our microbenchmark suite specified in NxKernel measures the effective NeuroCore barrier synchronization time, the effective NeuroCore peak operation rate for the chief compute operations SynOps, SynMem reads, and DendOps, and the effective peak bandwidth for NoC mesh links.
Measuring effective rates without delving into circuit-level details allows our microbenchmark suite to yield an easily understood and useful characterization of compute and communication.

\subsubsection{Microbenchmark Preliminaries}
For all microbenchmarks, we use either dense or sparse synaptic weight matrix encoding on Loihi 2.
Dense encoding stores all synaptic weights---even zero-valued weights---and sparse encoding stores only non-zero weights \cite{golub2013matrix}. We take advantage of these encoding schemes' differences in memory requirements and Loihi 2's automatic skipping of SynOps for zero-valued synaptic weights to stress different operations in different microbenchmarks.

For all microbenchmarks, we measure runtime for 100 timesteps on Loihi 2, record the execution time using the runtime profiler, and compute the mean time per step to infer effective computation or communication rates.

\subsubsection{Barrier Synchronization Microbenchmark}
Effective barrier synchronization time is measured with a dummy network in which one neuron in an origin NeuroCore is connected to another neuron on a destination NeuroCore through an identity synapse (\fig\ref{fig:microbenchmarks_1}a).
However, the neuron on the origin core is non-spiking, so it never emits a spike, and the neuron on the destination core never receives a spike. 
So every algorithmic timestep, only 1 DendOp (very rapidly) occurs in parallel on both NeuroCores, and the rest of execution time is solely due to synchronization time between algorithmic timesteps.
Thus measuring the mean time per step for this workload yields the effective barrier synchronization time.

\subsubsection{DendOp Microbenchmark}
Effective DendOp time is measured with a single-NeuroCore workload with $N$ non-spiking neurons (\fig\ref{fig:microbenchmarks_1}b).
Each algorithmic timestep, each neuron is trivially updated (no update); $N$ DendOps and barrier synchronization occur. 
Importantly, for large $N$, the DendOp time dominates; see \fig\ref{fig:microbenchmarks_measurements}a.
Thus we run this workload with $N=4095$ and divide measured mean time per step by $N$ to yield the effective DendOp time.

\subsubsection{SynOp Microbenchmark}
Effective SynOp time is measured with a linear layer: an origin core with $N$ sender neurons sends spikes to a destination core with $N$ neurons receiving through an $N\times N$ synaptic weight matrix of all ones (\fig\ref{fig:microbenchmarks_1}c).
On each timestep, each sender neuron emits a spike, and the spike's graded value is multiplied by $N$ synaptic weights at the destination core: one SynOp per destination neuron.
These products are accumulated for every neuron on the destination core across all $N$ incoming spikes, thus $N^2$ SynOps occur in total.
The destination neurons are identity neurons, so they re-emit the result of these accumulations as graded spikes.
However, these neurons do not have outgoing connections, so no spikes are actually emitted.
In summary, the time per step has contributions from $N$ DendOps, $N^2$ SynOps, and barrier synchronization, and for large $N$, the $N^2$ SynOps dominate (\fig\ref{fig:microbenchmarks_measurements}b); we divide the mean time per step for large $N$ by $N^2$ to calculate the effective SynOp time.
Importantly, we use dense-packed 1-bit synaptic weights, as this maximizes the number of synaptic weights per byte of synaptic memory; this minimizes SynMem reads while maximizing SynOps. 

\subsubsection{SynMem Read Microbenchmark}
Effective SynMem read time is measured with the same workload as in the SynOp microbenchmark, but with a dense-packed identity synaptic weight matrix with 8-bit weights (\fig\ref{fig:microbenchmarks_1}d).
Since the synaptic weight matrix is dense-packed, the destination NeuroCore must read all synaptic weight entries, including all zero entries. 
Hence, for the $N\times N$ identity matrix with large $N$, many SynMem reads occur; however, most synaptic weights read are zero, so few SynOps occur because Loihi 2 automatically skips SynOps for zero-valued weights. 
Moreover, we further exaggerate the SynMem read to SynOp ratio by using 8-bit weights, the maximum weight size on Loihi 2.
Thus for large $N$, time per step is dominated by SynMem reads, and we divide the mean time per step by the number of SynMem reads per step to measure effective SynMem read time (\fig\ref{fig:microbenchmarks_measurements}c).

\subsubsection{Link Bandwidth Microbenchmark}
Effective NoC link bandwidth is measured with a multi-core workload: 
$M$ origin-destination NeuroCore pairs (\fig\ref{fig:microbenchmarks_1}e) communicate over a common mesh link (\fig\ref{fig:microbenchmarks_2}a).
In each pair, $N$ neurons on the origin core spike every timestep, sending to $N$ identity neurons on the destination core through a sparse-packed identity weight matrix synapse.
Since the identity synaptic weight matrix has few non-zero entries and is sparse-packed, few SynMem reads and few SynOps occur during each timestep---on the order of $N$ SynMem reads and SynOps.
Meanwhile, $N$ spike messages travel from the origin core to the destination core, and as we arrange the cores as in \fig\ref{fig:microbenchmarks_2}a, all traffic from the origin cores coalesces along the vertical NoC links in accordance with dimension-order routing.
This stresses the NoC links, especially as the number of neurons $N$ and number of origin-destination NeuroCore pairs $M$ grows; the implied  link bandwidth is the number of bits transferred over the link during each timestep (number of bits per message $\times N \times M$) divided by the mean timestep duration (\fig\ref{fig:microbenchmarks_measurements}d).
We run this workload with $N=4095$ neurons and $M=12$ origin-destination group pairs to measure effective link bandwidth. 

\section{Max-affine runtime model}

Given Loihi 2's asynchronous architecture, time spent by NeuroCores and the NoC performing computations and communications naturally overlap, giving rise to a simple lower-bound max-affine runtime model.
Given that NeuroCores synchronize between algorithmic timesteps, we estimate the SynOp compute time per timestep as the maximum SynOps performed by any NeuroCore (the heaviest-loaded one) multiplied by the effective time per SynOp measured from the microbenchmark suite.
Analogously, we estimate how long the NeuroCores take to perform SynMem reads and DendOps.
To estimate the time the NoC takes to deliver all communication, we take the heaviest link load from TrafficStats and multiply it by the inverse link bandwidth from the microbenchmark suite.
Assuming overlap in computations and communications and adding  minimum barrier synchronization time, we estimate the time per step as
\begin{align} 
T &= \max \left( N_{\text{DO}} T_{\text{DO}},  N_{\text{SO}} T_{\text{SO}} ,  N_{\text{SMR}} T_{\text{SMR}}, N_{\text{LL}}/B, T_{\text{BS}} \right)
\label{eq:max_affine_model}
\end{align}
where $N_{\text{DO}}$, $N_{\text{SO}}$, and $N_{\text{SMR}}$ are the highest numbers of DendOps, SynOps, and SynMem reads, respectively, among all NeuroCores per timestep, $N_{\text{LL}}$ is the number of bits transferred over the heaviest-loaded NoC link per timestep, $T_{\text{DO}}$, $T_{\text{SO}}$, and $T_{\text{SMR}}$ are the effective time per DendOp, SynOp, and SynMem read, respectively, $B$ is the effective bandwidth of a NoC link, and $T_{\text{BS}}$ is the effective barrier synchronization time.
Notably, $T$ is a lower-bound estimate, as computations and communications do not overlap perfectly in reality, and there are other computation stages and operations on Loihi 2 that we do not account for explicitly in this model.

\section{Loihi 2 Results}

For several workloads and configurations, we measure time per step on Loihi 2 and compare it to our max-affine model's predicted time per step.
We find that the max-affine model indeed reflects a lower-bound estimate for runtime, and moreover, can be a near-exact runtime predictor.
This close correspondence  holds across compute and communication bottlenecks.
Furthermore, with a communication-bottlenecked workload, we explore Loihi 2 spatial placement options, which significantly affect mesh traffic, and show that our max-affine model's predictive power holds across this space.

\begin{figure}[h]
\centerline{\includegraphics[width=1\columnwidth]{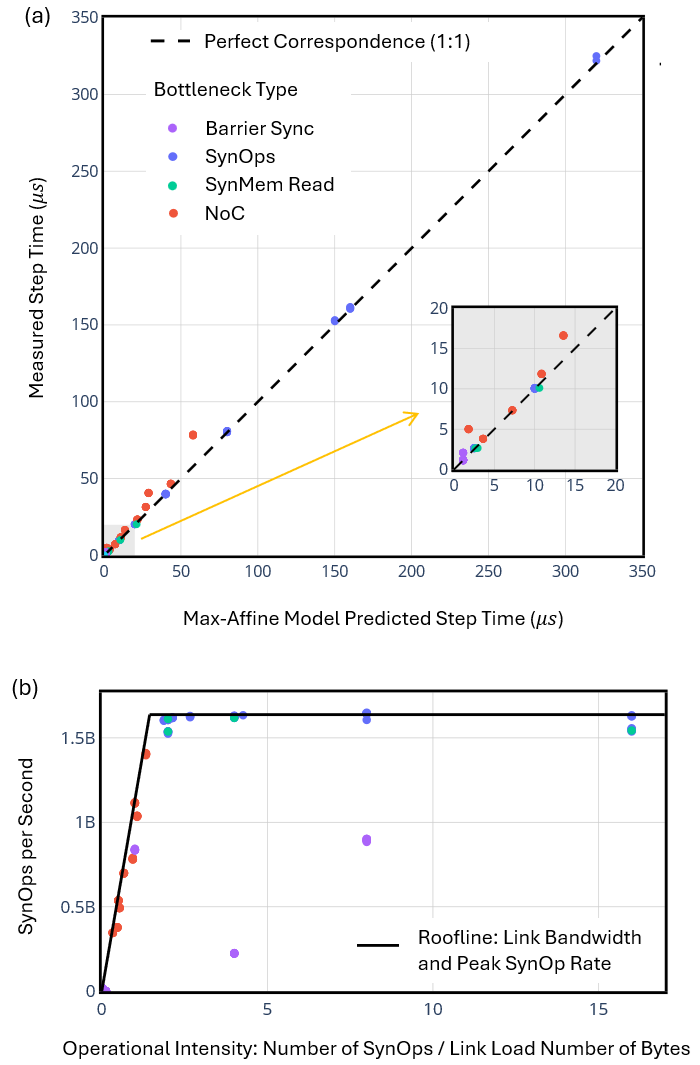}}
\vspace{-1mm}
\caption{Correspondence between modeled and measured performance for a dense linear layer.
(a) Correspondence over many dense linear layer configurations. The close correspondence across different bottleneck types (colors) provides evidence of the max-affine model's predictive power; the Pearson correlation coefficient is 0.996.
(b) Roofline visualization for the dense linear layer over many configurations. NoC-bottlenecked configurations trace the roofline's sloped portion and SynOp-bottlenecked configurations trace the roofline's horizontal portion. All other points lie under the roofline. 
Please note that throughout this work error bars are omitted in linear layer workloads as over 3 trials, the measured mean time per step varied by at most 2\%, and the microbenchmark system characteristic measurements underlying the max-affine model varied by at most 0.5\%.
\loihiog
\vspace{-2mm}
}
\label{fig:dense_linear}
\end{figure}

\subsubsection{Dense Linear Layer}
For a dense linear layer workload at a variety of scales and configurations, we observe a correspondence between measured and estimated time per step (\fig\ref{fig:dense_linear}a). 
The dense linear layer workload is the same as the SynOp microbenchmark workload (\fig\ref{fig:microbenchmarks_1}c), except the number of neurons $N$ and the $N\times N$ all-ones synaptic weight matrix can be much larger, with the origin and destination neuron populations distributed among many cores.
We configure the workload over 1, 4, 8, 16, 32, or 60 origin and destination NeuroCores (up to 120 cores in total), 1, 16, 32, 64, 128, or 256 neurons per core (the number of origin cores times the number of neurons per core yields the number of neurons $N$), and 1, 2, 3, ..., or 8 bit synaptic weights; we place all origin cores in the bottom half of the Loihi 2 chip and all destination cores in the top half, similar to \fig\ref{fig:microbenchmarks_2}a.
We include all configurations in this product space that fit in NeuroCore memory in \fig\ref{fig:dense_linear}a.
Furthermore, we observe the roofline nature of the max-affine  model (\fig\ref{fig:dense_linear}b).

\begin{figure}[h]
\centerline{\includegraphics[width=1\columnwidth]{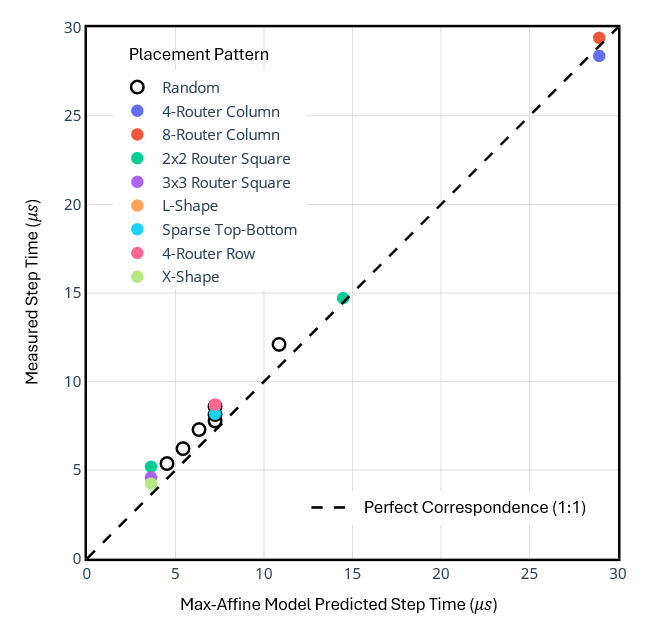}}
\vspace{-3mm}
\caption{
Correspondence between modeled and measured performance for a traffic-dominated tiled identity workload. For the same traffic-dominated workload mapped to the same number of NeuroCores, different placement patterns (symbols/colors) of the NeuroCores vastly alter the runtime as different placement patterns yield different levels of NoC traffic congestion. Over all placement patterns, however, the max-affine model maintains predictive power; the Pearson correlation coefficient is 0.999.
\loihiog
}
\label{fig:tiled_identity}
\end{figure}

\begin{figure}[h]
\centerline{\includegraphics[width=1\columnwidth]{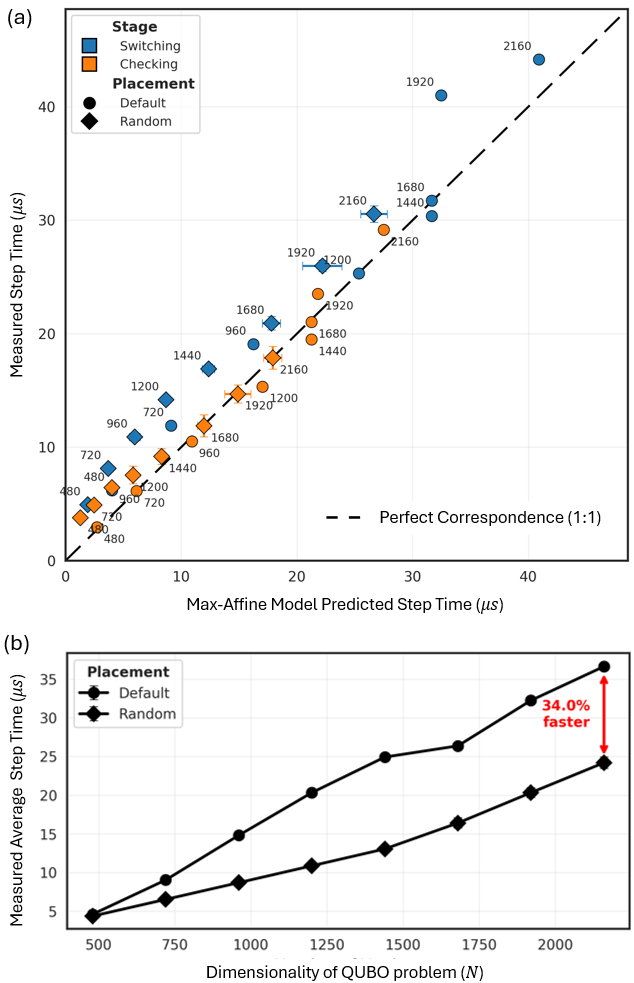}}
\vspace{-2mm}
\caption{QUBO solver runtime measurements. (a) Correspondence between measured time per step and max-affine model estimate for switching and checking stages of the QUBO solver under different placement patterns and different QUBO problem dimensionality (numbers shown near datapoints). The default placement pattern fills the chip column-by-column, starting from the bottom left core, measurements are averaged over 250 time steps and three trials, and error bars show one standard deviation.
The Pearson correlation coefficient is 0.970.
(b) Runtime advantage of random placement over default placement. Random placement runs faster because it more uniformly distributes traffic, especially for larger QUBO problems.
\loihiqubo
\vspace{-4mm}
}
\label{fig:qubo}
\end{figure}

\subsubsection{Traffic-bottlenecked Linear Layer}
To evaluate the runtime model under conditions where communication time dominates and spatial NeuroCore placement choices become important, we run a linear layer workload with a sparse tiled-identity synaptic weight matrix under a variety of placement configurations.
We again observe a correspondence between the max-affine model estimated step time and measured step time on Loihi 2; see \fig\ref{fig:tiled_identity}.
The tiled identity workload has 8 origin cores sending to 8 destination cores with 1024 neurons per core.
The connectivity matrix is $8192\times 8192$, composed of $1024\times 1024$ identity matrix tiles connecting each origin core's $1024$ neurons to all neurons in a destination core.
The $8$ $1024\times 1024$ identity matrix tiles stored on each destination core are sparse-packed, taking little memory.
This sparse-packed tiled-identity connectivity ensures that all neurons on all origin cores are connected to all neurons on all destination cores, maximizing NoC traffic, while keeping SynOps and SynMem reads at a minimum.
We run this workload under several different placement patterns, including random placement and an X-shaped placement; see \fig\ref{fig:microbenchmarks_2}bc for examples.
Crucially, the placement patterns affect the traffic routes from the origin cores to the destination cores---some patterns cause more congestion than others---and this is reflected in our max-affine performance model (\eq\ref{eq:max_affine_model}) as a change in the heaviest link load, $N_{LL}$.
The X-shaped pattern exhibited the lowest time per step.

\subsubsection{QUBO Solver}
We further demonstrate the predictive power of our max-affine performance model by estimating the runtime of a Quadratic Unconstrained Binary Optimization (QUBO) combinatorial solver \cite{pierro2024solving, US20250165765A1} (\fig\ref{fig:qubo}).
In the QUBO solver, a QUBO problem $\min_x x^T Q x$ where $x\in{\{0,1\}}^N$ is encoded in the states and recurrent connectivity of a population of $N$ neurons; the $i$th neuron encodes $x_i$.
An additional temperature neuron and loss neuron are connected to this population to facilitate annealing and progress tracking, but the bulk of compute and communication are performed by the population of $N$ neurons. 
As the QUBO solver runs, each of the $N$ neurons alternates between two stages: checking and switching. In the checking stage, the neurons perform a calculation and coordinate to determine an appropriate probability with which to flip their state variable $x_i$ (from 0 to 1, or vice versa). In the switching stage, the neurons stochastically flip their state variable $x_i$ according to the previously computed probability and communicate updates to other neurons so that the checking stage can be repeated with an up-to-date state of the QUBO problem.
Alternating checking and switching stages continue until the QUBO loss converges to a satisfactory level.

The QUBO solver is a traffic-bottlenecked workload, and we can analyze the checking stage and switching stage separately to form an estimate of the mean time per step in each stage under a variety of QUBO problem sizes and NeuroCore placement patterns; see \fig\ref{fig:qubo}.
Importantly, the traffic pattern is the same for both the checking stage and the switching stage, as they use the same recurrent connectivity structure of the QUBO matrix $Q$.
Thus, we can use the TrafficStats tool on the $N$ neurons and their recurrent connectivity to determine the heaviest link load. 
However, importantly, the TrafficStats tool by default assumes that all neurons emit spikes every timestep, but in the QUBO solver, neurons often do not spike, as they spike stochastically.
We account for this by measuring the activity rate $(1 - \text{activity sparsity})$ in the checking stage and switching stage, and multiplying the activity rate by the heaviest link load from TrafficStats.
This yields an estimate that corresponds to measured mean time per step for the checking stage and switching stage over a variety of QUBO problem sizes (\fig\ref{fig:qubo}).

\section{Analytical model for NoC congestion}

We can gain further insight into how different placement patterns of a linear layer workload congest the mesh to different degrees by deriving analytical expressions for the heaviest link load given a placement pattern.
For simplicity, we conceptually aggregate the two-mesh bidirectional links on Loihi 2 into effective single-mesh bidirectional links as traffic is typically distributed across both meshes, and we focus on variations of the highest-performing placement pattern, the X-shaped placement.
Namely, we consider placement patterns in which each pair of origin and destination cores is assigned to a router and all routers have at most two populated NeuroCores, like in \fig\ref{fig:microbenchmarks_2}c.
Intuitively, assigning an origin core and destination core to the same router keeps at least one communication path off the router-to-router mesh links (the communication path between the origin core and the destination core on the same router) and also minimizes any single router's burden for the total  traffic coming into each destination core (each destination core is a sink, and having more than one destination core on a single router doubles the traffic incoming to that router).
Importantly, we will see that the minimum possible heaviest link load is achievable under these conditions.

We calculate the heaviest link load for a linear layer by mathematically defining the placement pattern and tracing the consequences of all-to-all communication under dimension-order routing.
For a linear layer workload with $M$ origin and $M$ destination cores with all origin core neurons connected to all destination core neurons (such as the all-ones dense linear layer workload or the tiled identity linear layer workload) and the same number of neurons per core, $N$, on all cores, we express the placement pattern as a matrix $P_{ij} \in {\{0,1\}}^{n\times m}$ and $\sum_{i=1}^n \sum_{j=1}^m P_{ij} = M$, where $1$ indicates that a router at X-Y position $(i,j)$ is populated with an origin-destination core pair and $0$ indicates an unpopulated router.
Notably, we are limited to $n=8$ and $m=4$ on a single Loihi 2 chip, but we can nonetheless explore scaling beyond these limits in theory.
For the sake of brevity, we derive only the heaviest leftward router-to-router link load, as by symmetry this expresses heaviest rightward router-to-router link load, and similar calculations can be performed for upward and downward router-to-router links with analogous results.
Furthermore, we quantify traffic in units of $N$ messages, as each origin core is sending $N$ messages to each of the $M$ destination cores, i.e., we will drop the factor of $N$ neurons per core in expressing link loads.
Notably, this means that the link load is $M$ for all router-to-core and core-to-router links; this sets a floor for calculations of the heaviest link load from among all NoC links.

We calculate the heaviest leftward router-to-router link load by deriving an expression for the leftward link load at any router taking into account dimension-order routing of messages from all origin cores.
Let us first consider the routing of messages from a populated router ($P_{ij}=1$) at coordinates $(i,j)$; note that we consider $i=1$ the top row of routers and $j=1$ the leftmost column of routers, thus leftward-directed traffic is from higher values of $j$ to lower values of $j$.
This router has one origin core, which emits $M$ messages. Each message is destined for a different destination core and its associated router.
Because messages are routed horizontally first, we know that the number of messages that route horizontally and then route vertically up and down column $l$ is $\text{colsum}_l = \sum_{k=1}^n P_{kl}$, i.e., all messages destined for a given column must arrive in that column via horizontal routing. 
Now, consider the leftmost router in a row $i$: $(i,1)$. There is no leftward traffic as it is on the edge of the NoC.
Next, consider the second leftmost router in row $i$: $(i, 2)$. $\text{colsum}_1$ remains to be routed vertically at the leftmost router $(i,1)$, so the leftward traffic from router $(i,2)$ is $\text{colsum}_1$.
Next, consider the third leftmost router in row $i$: $(i, 3)$. $\text{colsum}_1 + \text{colsum}_2 $ remains to be routed either vertically at $(i,2)$, or $(i,1)$, thus the leftward traffic from router $(i,3) $ is $\text{colsum}_1 + \text{colsum}_2 $.
This is a cumulative sum, and this pattern continues until $l = j$; at $l = j +1$, traffic is now traveling to the right as the origin router is in column $j$, and thus leftward traffic at node $(i,j+1)$ is zero, and similarly for all coordinates further right.
We write this cumulative sum for the leftward traffic at $(i,l)$ from the router $(i,j)$, take a sum over all $j$ to get the total leftward traffic from all routers in row $i$, and take the max over all leftward links $(i,l)$ to obtain the heaviest leftward router-to-router link load:
\begin{align}
N_{\text{LL}}^{\text{(router-to-router)}} &= \max_{i,l} \left( \sum_{j=l}^m P_{ij} \sum_{l'=1}^{l-1} \text{colsum}_{l'} \right).
\label{eq:leftward}
\end{align}

With \eq\ref{eq:leftward}, we can plug in particular placement patterns and explore how the heaviest router-to-router link load grows with the hypothetical size of the NoC. 
For a saturated rectangle, $P_{ij}=1 \; \forall 1 \le i \le n, 1 \le j \le m$, the heaviest leftward link load is $N_{LL}=nm^2/4$.
A similar calculation for upward link load yields a symmetric formula in $n$ and $m$, and this implies the intuitive result that the optimal rectangular placement is a square with $n=m$; $n\neq m$ leaves room to reshape the rectangle to achieve a lower heaviest link load.
For the square, the number of origin-destination core pairs is $M=\sum_{i,j} P_{ij}=n^2$, and so the heaviest link load is $\frac{M^{3/2}}{4}$.
In contrast, an X-shaped placement $P_{ij} = \delta_{ij} + \delta_{i, n-j+1} \, \forall i,j \in \{1,...,n\}$, where $n$ is evenly divisible by $2$, yields $N_{LL} = 2(n - 1)$. In the X-shaped placement, $M = 2n$, and thus the heaviest router-to-router link load is $M-2$.
As noted earlier, the core-to-router link load is already $M$ for all placement patterns, thus the X-shape's $M-2$ heaviest router-to-router link load achieves the minimum possible heaviest link load overall.

\begin{figure}[h]
\vspace{-0.1mm}
\centerline{\includegraphics[width=1\columnwidth]{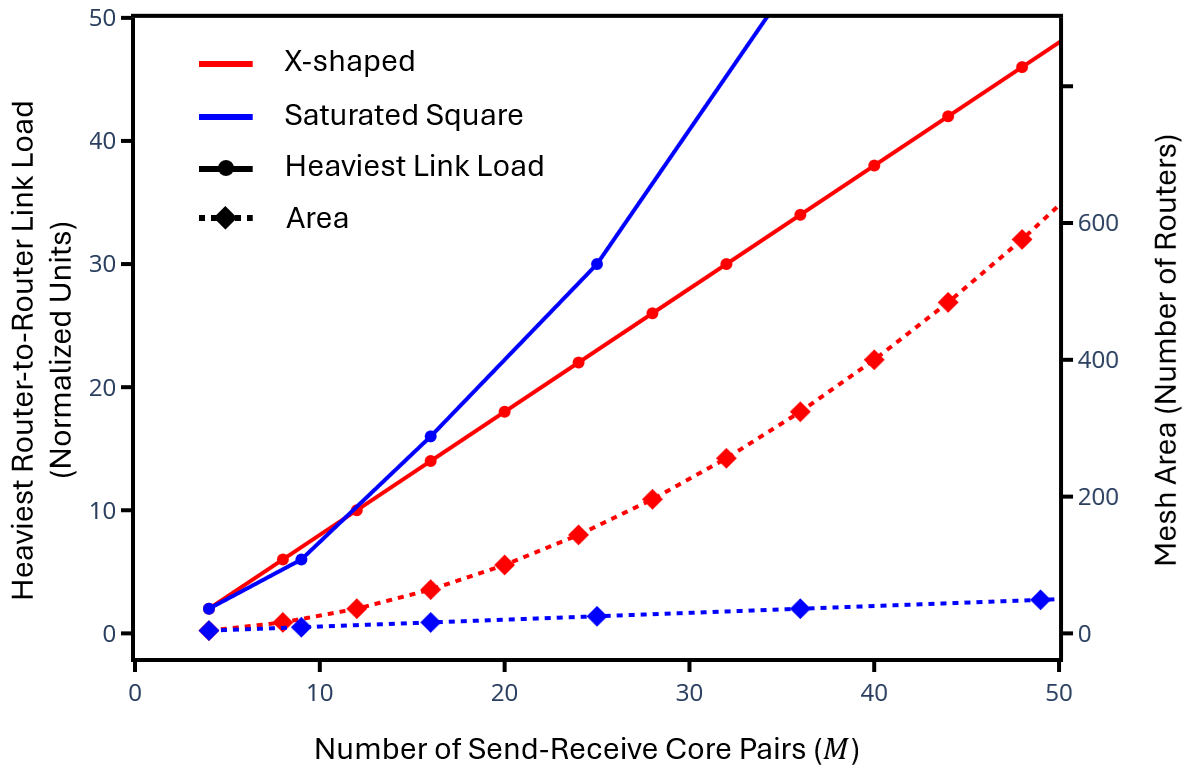}}
\vspace{-2mm}
\caption{Heaviest router-to-router link load and NoC area as the number of processors scales for different placement patterns. A tradeoff exists in placement pattern compactness and speed: runtime grows linearly and area grows superlinearly for the X-shaped placement pattern, and vice versa for the saturated square placement pattern. 
\vspace{-2.5mm}
}
\label{fig:scaling}
\end{figure}

Importantly, there is a scaling difference between the saturated square placement and the X-shaped placement: the saturated square grows superlinearly as $M^{3/2}$, while the X-shape grows linearly in $M$. 
Thus different placement patterns give rise to a tradeoff between mesh size and link load: for the same size of linear layer ($M$), the X-shape has lower heaviest link load but larger NoC area, while the saturated square has heavier link load but smaller NoC area (\fig\ref{fig:scaling}).

\eq\ref{eq:leftward} applied to other placement patterns, such as an identity placement $P_{ij} = \delta_{ij}$, yields $N_{LL}=M -1$, or for a generalized permutation matrix placement in which $\sum_j P_{ij} = a \; \forall i$ and $\sum_i P_{ij} = a \; \forall j$ for some constant $a$, a bound can be derived that shows $N_{LL} \le a M$. This illustrates that there exists a variety of placement patterns with linear growth in $M$ with different constant factors.

\section{Discussion and future work}

This work leveraged the profiling tools on Loihi 2 to introduce a max-affine performance model for runtime, validated the model on a linear layer kernel (matrix-vector multiplication) and a QUBO solver, and studied the communication time ramifications and scalings of different spatial placement patterns.
Our work can help researchers design performant kernels on Loihi 2 and may help inspire performance models for other neuromorphic hardware platforms.
A performance model for Loihi 2 including energy in addition to runtime and validated using Loihi 2's power profiling capabilities is forthcoming in a future publication.

Importantly, even though Loihi 2 operates asynchronously, we demonstrate the relatively simplicity with which runtime can be modeled for a fundamental and ubiquitous kernel---matrix-vector multiplication---and we treat communication in sufficient detail so that the model is predictive of actual communication time. 
This is especially important for modeling runtime in latency-critical application domains such as robotics, as spreading out workloads across many neuromorphic compute cores to achieve high performance and break the memory-bandwidth wall is a primary advantage afforded by neuromorphic hardware.

There are several limitations that highlight opportunities for future research. 
This work focuses on a single linear layer and a QUBO solver; a larger variety of kernels and the composition of kernels will provide a more comprehensive assessment of the predictive power of this max-affine model. 
While exhibiting a tight correspondence with measured runtime, our model also exhibits deviations.
These modeling errors can be accommodated for by engineering built-in runtime tolerances in latency-critical applications, and the errors could also be reduced with a more complex runtime model that takes into account finer-grained and dynamic runtime effects beyond those considered in this work. 
This work directly applies to fall-through mode, where generally one kernel is executing on Loihi 2 at a time and each kernel has the full NoC bandwidth.
However, in pipeline mode, where several kernels can run in parallel, further congestion in the NoC can occur. (See \cite{abreu2025neuromorphic} for fall-through and pipeline mode.)
Researchers can use the TrafficStats tool to quantify multi-kernel NoC congestion.

\section*{Acknowledgments}
We thank Timothy Shea, Andreas Wild, Mike Davies, Andrew Lines, and Ruokun Liu for their helpful support.

\printbibliography

\end{document}